\documentclass{article}
\usepackage{amsmath,amsfonts,bm}
\usepackage{multirow}
\usepackage{threeparttable}
\usepackage{tablefootnote}
\usepackage{microtype}
\usepackage{graphicx}
\usepackage{subfigure}
\usepackage{booktabs} 
\usepackage[keeplastbox]{flushend}

\usepackage{hyperref}
\usepackage{cleveref}

\usepackage[accepted]{icml2020}

\icmltitlerunning{A Graph to Graphs Framework for Retrosynthesis Prediction}

\newcommand{\modelname}{G2Gs~}

\begin{document}

\twocolumn[
\icmltitle{A Graph to Graphs Framework for Retrosynthesis Prediction}



\icmlsetsymbol{equal}{*}

\begin{icmlauthorlist}
\icmlauthor{Chence Shi}{pku}
\icmlauthor{Minkai Xu}{sjtu}
\icmlauthor{Hongyu Guo}{nrc}
\icmlauthor{Ming Zhang}{pku}
\icmlauthor{Jian Tang}{mila,cifar,hec}

\end{icmlauthorlist}

\icmlaffiliation{pku}{Department of Computer Science, School of EECS, Peking University}
\icmlaffiliation{sjtu}{Shanghai Jiao Tong University}
\icmlaffiliation{nrc}{National Research Council Canada}
\icmlaffiliation{mila}{Montr\'eal Institute for Learning Algorithms (MILA)}
\icmlaffiliation{cifar}{Canadian Institute for Advanced Research (CIFAR)}

\icmlaffiliation{hec}{HEC Montr\'eal}

\icmlcorrespondingauthor{Chence Shi}{chenceshi@pku.edu.cn}
\icmlcorrespondingauthor{Jian Tang}{jian.tang@hec.ca}

\icmlkeywords{Machine Learning, ICML}

\vskip 0.3in
]



\printAffiliationsAndNotice{} 
\begin{abstract}

A fundamental problem in computational chemistry is to find a set of reactants to synthesize a target molecule, a.k.a. retrosynthesis prediction. Existing state-of-the-art methods rely on matching the target molecule with a large set of reaction templates, which are very computationally expensive and also suffer from the problem of coverage. In this paper, we propose a novel template-free approach called \modelname by transforming a target molecular graph into a set of reactant molecular graphs. \modelname first splits the target molecular graph into a set of synthons by identifying the reaction centers, and then translates the synthons to the final reactant graphs via a variational graph translation framework. Experimental results show that \modelname significantly outperforms existing template-free approaches by up to 63\%  
in terms of the top-1 accuracy 
and achieves a performance  close to that of 
state-of-the-art template-based approaches, 
but does not require domain knowledge and is much more scalable.
The code is available at \url{https://github.com/DeepGraphLearning/torchdrug}.

\end{abstract}
\section{Introduction}
Retrosynthesis, which devotes to find a set of reactants to synthesize a target molecule, is of crucial importance to the synthesis planning and drug discovery. The problem is challenging as the search space of all possible transformations is huge by nature. For decades, people have been seeking to aid retrosynthesis analysis with modern computer techniques~\citep{retro1969computer_aid}. Among them, machine learning plays a vital role and significant progress has been made recently~\citep{sara2016computer_retro, connor2018ml_retro}.

Existing machine learning works on retrosynthesis prediction mainly fall into two categories: template-based~\cite{Connor2017retrosim, Marwin2017neuralsym, Dai2019GLN} and template-free models~\cite{liu2017retro_seq2seq, pavel2019retro_transformer}. 
The template-based approaches  
match the target molecule with a large set of reaction templates, which define the subgraph patterns of a set of chemical reactions. For example,~\citep{Connor2017retrosim} proposed a similarity-based approach to select reaction templates for the target molecule.~\citep{Marwin2017neuralsym, Javier2019retro_multiscale_class} cast rule selection as a multi-class classification problem. Recently,~\citep{Dai2019GLN} treats chemistry knowledge as logic rules and directly models the joint probability of rules and reactants, which achieves the new state of the art.
Despite their great potential for synthesis planning, template-based methods, however, not only require expensive computation but also suffer from poor generalization on new target structures and reaction types.
Another line of research for retrosynthesis  prediction~\citep{liu2017retro_seq2seq, pavel2019retro_transformer} bypasses reaction templates and formulates retrosynthesis prediction as a sequence-to-sequence problem. Such approaches leverage the 
 recent advances in neural machine translation ~\citep{bahdanau2014nmt, vaswani2017transformer} and the  SMILES~\citep{weininger1988smiles} representation of molecules.
However, SMILES representation assumes a sequential order between the atoms in a molecule, which cannot effectively reflect the complex relationships between atoms in a molecule. 
As a result, these approaches fail to capture the rich chemical contexts and their interplays of molecules, resulting in  unsatisfactory  predictive performance.

To address the aforementioned issues, 
in this paper we represent each molecule as a graph and formulate retrosynthesis prediction as a graph-to-graphs translation problem. The so-called G2Gs leverages the powerful representation of graph for molecule and is a novel template-free approach, which is trained with an extensive collection of molecule reaction data.
Specifically, it consists of two key components: (1) a reaction center identification module, which splits synthons from the target molecule and reduces the one-to-many graph translation problem into multiple one-to-one  translation processes; (2) a variational graph translation module, which translates a synthon to a final reactant graph. 
We aim to model the probability of reactant graph $G$ conditioned on the synthon $S$, \textit{i.e.}, $P(G|S)$. As a synthon could be potentially translated to different reactants in different reaction contexts, a latent code $z$ is introduced to handle the uncertainty of reactant prediction, \textit{i.e.}, $P(G|z,S)$. Following existing work on graph generation~\cite{you2018graph, liu2018constrained, shi*2020graphaf}, we formulate reactant graph generation as a sequential decision process, more specifically, sequentially generating nodes and edges. The whole graph translation process can be efficiently and effectively optimized with the variational inference approach~\citep{kingma2014vae}. 


We evaluate our model on the benchmark data set USPTO-50k derived from a patent database~\citep{daniel2012phdthesis}, and compare it with both template-based and template-free approaches.
We experimentally show that \modelname significantly outperforms existing template-free baselines by up to 63\% in terms of the top-1 accuracy. These numbers are also approaching those obtained by the state-of-the-art template-based strategies, but our method excludes the need for domain knowledge and scales well to larger data sets, making it particularly attractive in practice.

\section{Related Work}

\textbf{Retrosynthesis Prediction }
Prior works on retrosynthesis prediction are primarily based on reaction templates, which are either hand-crafted by human experts~\citep{Markus2011reaction_template, sara2016computer_retro} or extracted from large chemical databases automatically~\citep{connor2017pred_react}.
Since there are hundreds of qualified templates for a target molecule by subgraph matching, selecting templates that lead to chemically feasible reactions remains a crucial challenge for these approaches.
To cope with such challenge,  \citep{Connor2017retrosim} proposed to select templates based on similar reactions in the data set.~\citep{Marwin2017neuralsym, Javier2019retro_multiscale_class} further employed neural models for rule selection. The state-of-the-art method~\citep{Dai2019GLN} 
leveraged the idea that 
the templates and reactants are hierarchically sampled from their conditional joint distributions. Such template-based approaches, however, still  suffer from poor generalization on unseen structures and reaction types, and the computational cost for subgraph isomorphism in these methods is often prohibitively expensive.

To overcome the limitation of template-based methods, template-free approaches have recently been actively investigated.    ~\citep{liu2017retro_seq2seq, pavel2019retro_transformer} formulated the task as a sequence-to-sequence problem on SMILES representation of molecules, and the same idea was adopted in~\citep{philippe2017transformer, Schwaller2018MolecularTF} for its dual problem, \textit{i.e.}, organic reaction prediction. However, these approaches are apt to ignore the rich chemical contexts  contained in the graph structures of molecules, and the validity of the generated SMILES strings can not be ensured, resulting in unsatisfactory performance. 
In contrast, our \modelname framework, directly operating on the graph structures of molecules, is able to generate 100\% chemically valid predictions with high accuracy, while excludes the need for  reaction templates and computationally expensive graph isomorphism.

Our work is also related to the chemical reaction prediction method presented in~\citep{jin2017weisfeiler} in terms of learning a neural model for reaction center identification. Nevertheless, both the  definition of the reaction center and the targeted task, namely  retrosynthesis prediction, 
distinguish our strategy from their algorithm. 


\textbf{Molecular Graph Generation }
Various deep generative models for molecular graph generation have recently been introduced~\cite{segler2017generating,olivecrona2017molecular,samanta2018designing,neil2018exploring,you2018graph, liu2018constrained, shi*2020graphaf}. 
Our \modelname framework is closely related to the state-of-the-art models that decompose the generation process into a sequence of graph transformation steps, including~\cite{you2018graph, liu2018constrained, shi*2020graphaf}. In these approaches, the generation procedure is formulated as a sequential decision making process by dynamically adding new nodes and edges based on current subgraph structures. For example,  \cite{you2018graph} introduced a reinforce policy network for the decision making.  \cite{shi*2020graphaf} presented an alternative sequential generation algorithm based on autoregressive flow called GraphAF. \cite{liu2018constrained} combined sequential generation with a variational autoencoder to enable continuous graph optimization in the latent space.

Unlike the aforementioned approaches, we here leverage  graph generation for retrosynthesis prediction, where novel components have to be devised for this specific task. For example, in the retrosynthesis scenario, given a specified product molecule, the reactants could be slightly different with diverse reaction conditions such as reagent and temperature. To cope with such intrinsic  multi-modality problem, in \modelname we also explicitly introduce low-dimensional latent variables to modulate the sequential graph translation process, aiming at  capturing the diverse output distributions.

\begin{figure}[tbhp]
	\centering
    \includegraphics[width=.79\linewidth]{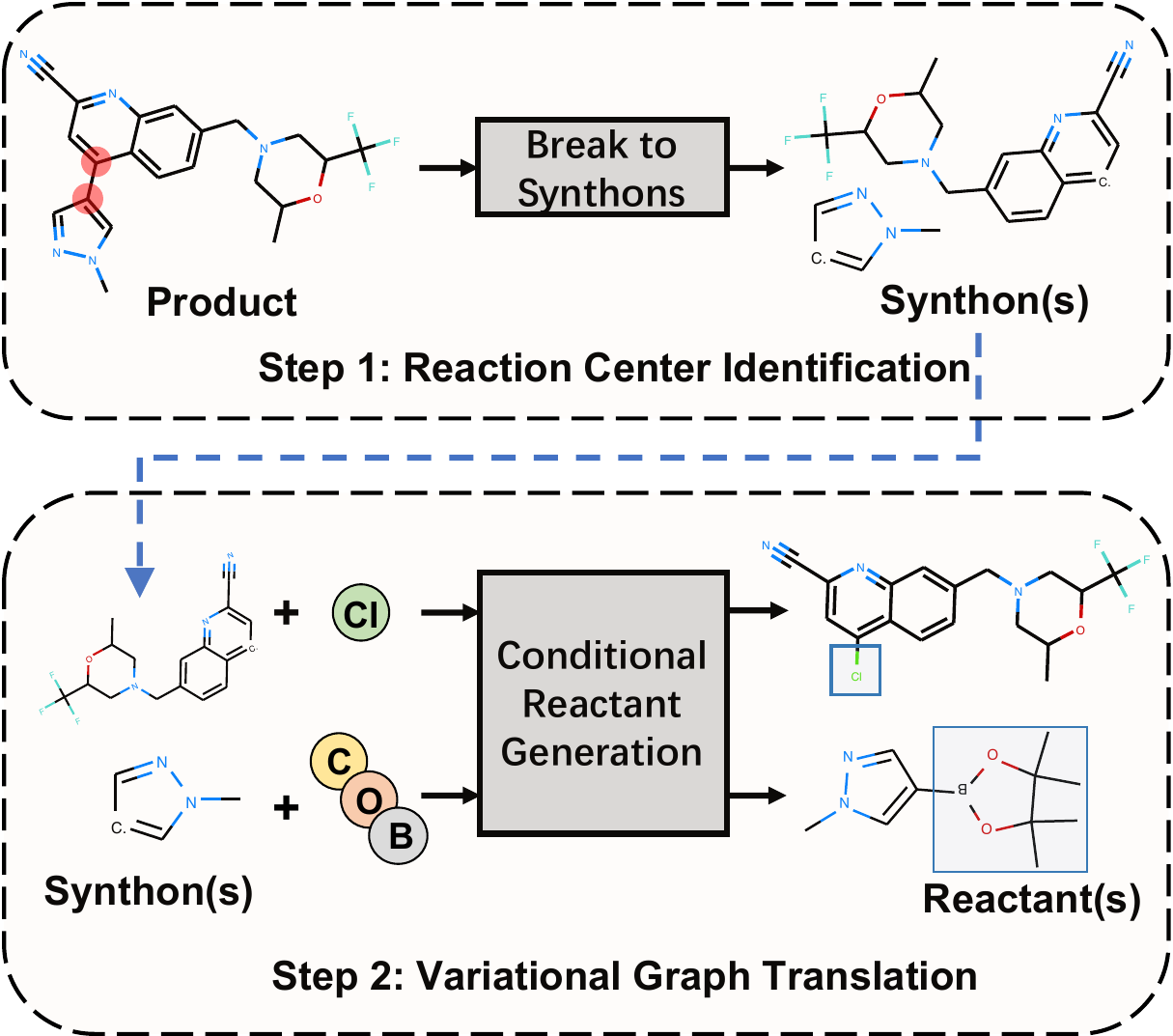}
    \caption{The overall framework of the proposed method. The reaction center identified by \modelname is marked in red. The product graph  is first split into synthons by disconnecting the reaction center. Based on resulted synthons, reactants are then generated via a series of graph transformations. The generated molecule scaffolds are bounded by blue bounding box.}
    \label{fig:framework}
    \vspace{-7pt}
\end{figure}

\section{The Graph to Graphs Framework}\label{sec:model}

In this paper, we formulate the retrosynthesis task as a one-to-many graph-to-graphs translation problem. 
In specific, we first employ a graph neural network to estimate the reactivity score of all atom pairs of the product graph, and the atom pair with the highest reactivity score above a threshold will be selected as the reaction center. We then split the product graph into synthons by disconnecting the bonds of the reaction center resulted. Finally, basing on the obtained synthons, the reactants are generated via a series of graph transformations, where a latent vector is employed to encourage the model to capture the transformation uncertainty and generate diverse predictions.
The proposed framework is illustrated in Figure~\ref{fig:framework}, and will be  discussed in  detail next. 

\subsection{Preliminaries}
\label{subsec:nation_notation}

\textbf{Notation }
In this paper, a molecule is represented as a labeled graph $G = (A, X)$, where $A$ is the adjacency matrix and $X$ the matrix of node features. Let the number of atoms be $n$, the number of bond types be $b$, and the dimension of node features be $d$, then we have $A \in \{0, 1\}^{n \times n \times b}$ and $X \in \{0, 1\}^{n \times d}$. $A_{ijk}=1$ here indicates a bond with type $k$ between the $i^{th}$ and $j^{th}$ nodes.

\textbf{Retrosynthesis Prediction }
Formally, a chemical reaction can be represented as a pair of two sets of molecules $(\{G_i\}_{i=1}^{N_1}, \{G_j\}_{j=1}^{N_2})$, where $G_i$ denotes a reactant graph and $G_j$  a product graph. In retrosynthesis prediction, given the set of products $\{G_j\}_{j=1}^{N_2}$, the goal is to precisely predict the set of reactants $\{G_i\}_{i=1}^{N_1}$. Following existing work, in this paper we focus on the standard single-outcome reaction case, \textit{i.e.}, $N_2=1$, and thus simplifying  the  notation of a reaction $r$ as $(\{G_i\}_{i=1}^{N_1}, G_p)$. 

\textbf{Reaction Center and Synthon }\label{reaction_center}
Unlike that in~\citep{jin2017weisfeiler},  the phrase \textit{reaction center} here is used to represent an atom pair $(i, j)$ that satisfies two conditions: (1) there is a bond between the  $i^{th}$ and $j^{th}$ nodes in the product graph; (2) there is no  bond between the  $i^{th}$ and $j^{th}$ nodes in the reactant graph. 
Also, synthons are subgraphs extracted from the products by breaking the bonds in the reaction centers. These synthons can later be transformed into reactants, 
and a synthon may not be a valid molecule.

\textbf{Molecular Graph Representation Learning }
\label{subsec:rgcn}
Graph Convolutional Networks (GCN)~\citep{duvenaud2015convolutional,kearnes2016molecular,kipf2016semi,gilmer2017neural,schutt2017quantum} have achieved great success in representation learning for computational chemistry. Given a molecular graph $G=(A, X)$, we adopt a variant of Relational GCN (R-GCN)~\citep{schlichtkrull2018modeling} to compute both the node embeddings and the graph-level embedding. Let $k$ $\in \mathbb{R}$ be the embedding dimension and $H^l \in \mathbb{R}^{n\times k}$  the node embeddings at the $l^{th}$ layer computed by the R-GCN $(H^0 = X)$. $H^l_i$ represents the embedding of the $i^{th}$ atom. At each layer, the R-GCN calculates the embedding for each node by aggregating messages from  different edge types:
\begin{equation}
H^{l}=\operatorname{Agg}\left(\operatorname{ReLU}\big(\{E_{i} H^{l-1} W_{i}^{l}\} \big| i \in(1, \ldots, b)\big)\right)
\end{equation}
where $E_{i} = A_{[:,:,i]} + I$ denotes the adjacency matrix of the $i^{th}$ edge type with self-loop and $ W_{i}^{l}$ is the trainable weight matrix for the $i^{th}$ edge type. $\operatorname{Agg}(\cdot)$ denotes an aggregation function chosen from summation, averaging and concatenation. 
We stack $L$ R-GCN layers to compute the final node embeddings $H^L$. The entire graph-level embedding $h_G$ can also be calculated by applying a $\operatorname{Readout}(\cdot)$ function to $H^L$~\cite{hamilton2017inductive}, \textit{e.g.}, summation.

\subsection{Reaction Center Identification}\label{subsec:reaction_center}
Given a chemical reaction $(\{G_i\}_{i=1}^{N_1}, G_p)$, we first derive a binary label matrix $Y \in \{0, 1\}^{n \times n}$ for each atom pair (\textit{i.e.}, bond) in the product $G_p$, indicating the reaction centers as defined in Section~\ref{subsec:nation_notation}; $Y_{ij}=1$ here indicates that the bond between $i^{th}$ and $j^{th}$ node in $G_p$ is a reaction center. It is worth noting that, both the reactants and the product are \textit{atom-indexed}. That is, each atom in the reactant set is associated with a unique index. This property enables us to  identify the reaction centers by simply comparing each pair of atoms in the product $G_p$ with that in a reactant $G_i$, forming the binary label matrix $Y$. With such label matrix, the Reaction Center Identification procedure in G2Gs is then formulated as a binary link prediction task as follows.

We use a $L$-layer R-GCN defined in Section~\ref{subsec:rgcn} to compute the node embeddings $H^L$ and the entire graph embedding $h_{G_p}$ of product $G_p$:
\begin{equation}
H^{L} = \operatorname{R-GCN} (G_p), \  h_{G_p}= \operatorname{Readout} (H^{L}).
\end{equation}
To estimate the reactivity score of the atom pair $(i, j)$, the edge embedding $e_{ij}$ is formed by concatenating  the node embeddings of the $i^{th}$ and $j^{th}$ nodes as well as the one-hot bond type feature. In addition, since a $L$-layer R-GCN can only gather information within $L$ hops of the center node, but the reactivity of a reaction center may also be affected by the remote atoms, we also enable the edge embedding  $e_{ij}$ to take into account the graph embedding (\textit{i.e.}, global structure), so as to leverage the knowledge of the remote atoms. 
Formally, we have:
\begin{equation}
e_{ij} = H^L_i \ \Vert \ H^L_j \  \Vert \ A_{ij} \ \Vert \ h_{G_p}
\end{equation}
where $\Vert$ denotes the vector concatenation operation. $H_i^L \in \mathbb{R}^k$ denotes the $i^{th}$ row of $H^L$ and $A_{ij}=A_{[i,j,:]} \in \{0, 1\}^b$ is the edge type of the atom pair $(i, j)$ in $G_p$. 
The final reactivity score of the atom pair $(i,j)$ is calculated as:
\begin{equation}
s_{ij} = \sigma(m_r(e_{ij}))
\end{equation}
where $m_r$ is a feedforward network that maps $e_{ij}$ to a scalar, and $\sigma(\cdot)$ denotes the $\operatorname{Sigmoid}$ function. In the case that a certain reaction type is known during retrosynthesis analysis, we can also include this information by concatenating its embedding to the input of the feedforward network, \textit{i.e.}, $e_{ij}$.

For learning, the reaction center identification module can be optimized by maximizing the cross entropy of the binary label matrix $Y$ as follows:
\begin{equation}
    \mathcal{L}_1 = -\sum\limits_{r}\sum\limits_{i \neq j  } \lambda Y_{ij}\text{log}(s_{ij}) +  (1-Y_{ij})\text{log}(1-s_{ij})
\end{equation}
where $\lambda \in [1, + \infty)$ is a weighting hyper-parameter, with the following purpose. In practice, among all the atom pairs there are typically a few reaction centers. As a result, the output logit of the feedforward network $m_r$ is often extremely small. Through  the weighting  $\lambda$, we are able to  alleviate such issue caused by the imbalanced class distributions.

During  inference, we first calculate the reactivity scores of all atom pairs, and then select the highest one above a threshold as the reaction center. 
Alternatively, one can select 
the top-$k$ atom pairs with the highest  scores above a threshold as the reaction center. This scenario will  yield more diverse synthesis routes but with the cost of inference time.

After collecting reaction centers, we then disconnect the bonds of the reaction centers in $G_p$, and treat each connected subgraph in $G_p$ as a synthon. Note that, in the case that none of the reaction center is identified, the product $G_p$ itself will be considered as a synthon. Doing so, we can extract all the synthons from the product $G_p$, and then formulate a one-to-one graph translation procedure to translate each resulted synthon to a final reactant graph. We will discuss this translation process in detail next. 

\begin{figure*}[hbtp]
	\centering
    \includegraphics[width=.80\linewidth]{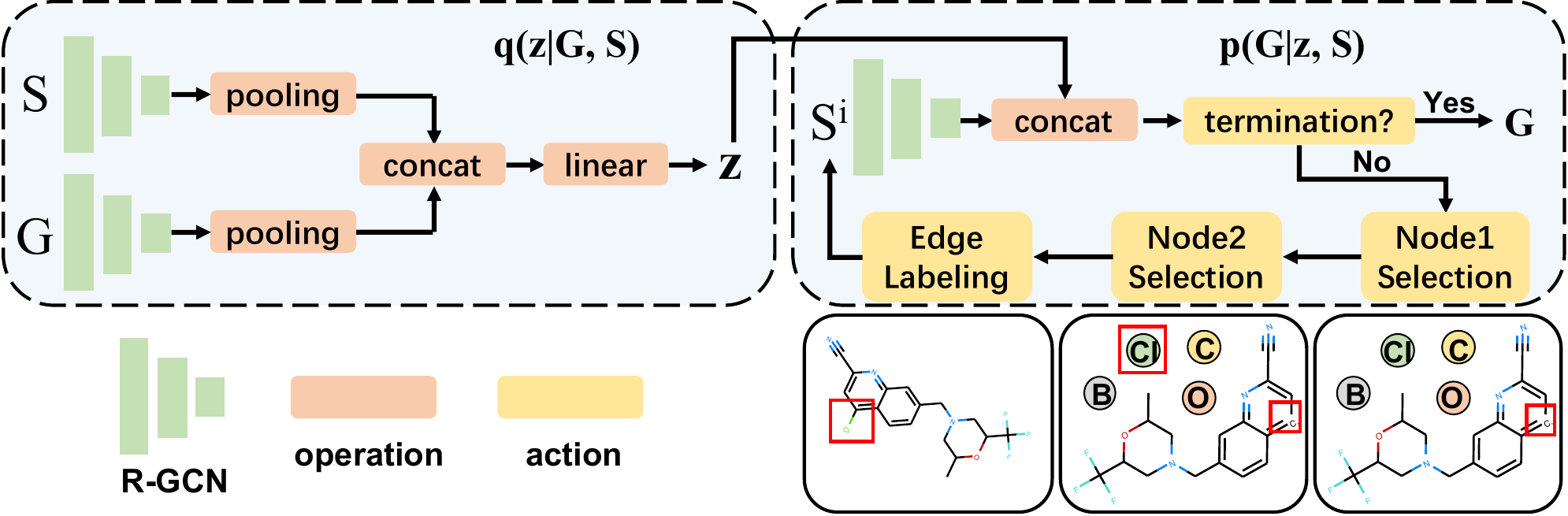}
    \caption{Illustration of the proposed variational graph translation module, including the encoder $q(z|G,S)$ (left) and the generative model $p(G|z,S)$ (right). The phases of the generation procedure are shown in the generative model. With a sampled latent vector $z$, the synthon graph $S$ enters a loop with nodes selection and edge labeling until the termination state is predicted.}
    \label{fig:syn2graph}
    \vspace{-7pt}
\end{figure*}

\subsection{Reactants Generation via Variational Graph Translation}
Given a chemical reaction $(\{G_i\}_{i=1}^{N_1}, G_p)$, we denote the set of synthons extracted from $G_p$ (as discussed in Section~\ref{subsec:reaction_center}) as $\{S_i\}_{i=1}^{N_1}$. For simplicity, we omit the subscript and denote a translation pair as $(S, G)$. 
In this setting, our goal  is to  learn a conditional generative model $p(G|S)$ that recovers the reactant molecule domain ($G$) from the synthon molecule domain ($S$). 
It is worth noting that, an intrinsic issue is that the same synthon can be translated to different reactants, which is known as the multi-modality problem.
In order to mitigate such multi-modality problem and encourage the module to model the diverse distribution of reactants, we incorporate a low-dimensional latent vector $z$ to capture the uncertainty for the graph translation process. Details are discussed next. 

\subsubsection{The Generative Model}\label{subsubsec:generative_model}
We build upon previous works on graph generation models~\cite{li2018learning,you2018graphrnn}. 
The generation of graph $G$ is conditioned on both the $S$ and the latent vector $z$.
In detail, we first autoregressively generate a sequence of graph transformation actions $(a_1, \cdots, a_T)$, and then apply them on the initial synthon graph $S$. Here $a_t$ is a one-step graph transformation (\textit{i.e.}, action) acting as a modification to the graph. Formally, let $\mathcal{T}$ be the collection of all trajectories $(a_1, \cdots, a_T)$ that can translate synthons $S$ to target reactants $G$ and $t \in \mathcal{T}$ be a possible trace. Then, we factorize  modeling $p(G|z,S)$ into modeling the joint distribution over the sequence of graph transformation actions $p(t | z, S)$. The connection between $p(G|z,S)$ and $p(t | z, S)$ is illustrated in Section~\ref{subsubsec:step2-learning}. With such a generative model, a synthon can be translated into a reactant by sampling action sequences from the distribution. Next, we describe the details of the generative procedure.

Let $S^{i}$ denote the graph after applying the sequence of actions $a_{1:i}$ to $S$, and $S^0 = S$. Then we have $p(S^i|S^{i-1},z)=p(a_{i}|S^{i-1},z)$. In previous graph generation models~\cite{li2018learning,you2018graphrnn}, each decision $a_i$ is conditioned on a full history of the generation sequence ($S^0, \cdots, S^{i-1}$), which leads to the stability and scalability problems arising from the optimization process. 
To alleviate these issues, 
our graph translation dynamics leverages the assumption of a Markov Decision Process (MDP), which satisfies
the Markov property that $p(S^i|S^{i-1},z)=p(S^i|S^{i-1},\cdots,S^0,z)$. The MDP formulation means that each action is only conditioned on the graph that has been modified so far. Hence the graph translation model $p(t | z, S)$ can be naturally factorized as follows:
\begin{equation}\label{eq:trace_fact}
p(t|z,S) = p(a_{1:T} | z, S) =
\prod\limits_{i=1}^T p(a_{i} | z, S^{i-1}).
\end{equation}

Before we detail the parameterization of distribution $P(a_{i} | z, S^{i-1})$, we formally introduce the definition of an action. An action $a_i$ is a tuple with four elements:
\begin{equation}
a_i = (a_{i}^1, a_{i}^2, a_{i}^3, a_{i}^4).
\end{equation}
Assuming the number of atom types is $m$, then $a_{i}^1 \in \{0, 1\}^2$ predicts the termination of the  graph translation procedure; $a_{i}^2 \in \{0, 1\}^n$ indicates the first node to focus; $a_{i}^3 \in \{0, 1\}^{n+m}$ indicates the second node to focus; $a_{i}^4 \in \{0, 1\}^b$ predicts the type of bond between two nodes. Then the distribution $p(a_{i} | z, S^{i-1})$ can be further decomposed into three parts: (1) termination prediction $p(a_{i}^1 | z, S^{i-1})$; (2) nodes selection $p(a_{i}^{2:3}, | z, S^{i-1}, a_{i}^1) $; (3) edge labeling $p(a_{i}^4 | z, S^{i-1}, a_{i}^{1:3})$.
We will discuss the parameterization of these components in detail next. 

\textbf{Termination Prediction }
Denote a $L$-layer R-GCN (Section~\ref{subsec:rgcn}) that can compute the node embeddings of an input graph as $\mathcal{R}(\cdot)$.
We parameterize the distribution as:
\begin{equation}\label{eq:term_pred}
\begin{aligned}
&H = \mathcal{R}(S^{i-1}), \  h_{S}= \operatorname{Readout} (H)  \\
&p(a_{i}^1 | z, S^{i-1}) = 
\tau\big(m_t(h_{S}, z)\big)
\end{aligned}
\end{equation}
where $\tau(\cdot)$ denotes the softmax function, and $m_t(\cdot)$ is a feedforward network. At each transformation step, we sample $a_{i}^1 \sim p(a_{i}^1 | z, S^{i-1})$. If $a_{i}^1$ indicates the termination, then the graph translation process will stop, and the  current graph $S^{i-1}$ is treated as the final reactant $G$ generated by the module.

\textbf{Nodes Selection }
Let the set of possible atoms (\textit{e.g.}, carbon, oxygen) to be added during graph translation be $\{v_1, \cdots, v_m\}$ and denote the collection as $V= \bigcup_{i=1}^m v_i$.
We first extend the graph $S^{i-1}$ to $\tilde{S}^{i-1}$ by adding isolated atoms, \textit{i.e.}, $\tilde{S}^{i-1} = S^{i-1} \bigcup V$.
The first node is selected from atoms in $S^{i-1}$, while the second node is selected from $\tilde{S}^{i-1}$ conditioned on the first node, by concatenating its embeddings with embeddings of each atom in $\tilde{S}^{i-1}$:
\begin{equation}\label{eq:node_select}
\begin{aligned}
&p(a_{i}^2 | z, S^{i-1}, a_{i}^1) = \tau
\big(
\beta_1 \odot
m_f(\mathcal{R}(\tilde{S}^{i-1}), z)
\big)\\
&a_{i}^2 \sim p(a_{i}^2 | z, S^{i-1}, a_{i}^1)\\
&p(a_{i}^3 | z, S^{i-1}, a_{i}^{1:2}) = \tau
\big(
\beta_2 \odot
m_s(\mathcal{R}(\tilde{S}^{i-1}), z, a_{i}^2)
\big)\\
&a_{i}^3 \sim p(a_{i}^3 | z, S^{i-1}, a_{i}^{1:2})\\
\end{aligned}
\end{equation}
where $m_f(\cdot)$ and $m_s(\cdot)$ are feedforward networks. $\beta_1$ and $\beta_2$ are masks to zero out the probability of certain atoms being selected. 
Specifically, $\beta_1$ forces the module to select node from $S^{i-1}$, and $\beta_2$ prevents the first node from being selected again.
In this case, only the second node can be selected from $V$, and it corresponds to adding a new atom to $S^{i-1}$.
The distribution of node selection $p(a_{i}^{2:3}, | z, S^{i-1}, a_{i}^1)$ is the product of the two conditional distributions  described above.

\textbf{Edge Labeling }
The distribution for edge labeling is parameterized as follows:
\begin{equation}\label{eq:edge_label}
\begin{aligned}
&p(a_{i}^4 | z, S^{i-1}, a_{i}^{1:3}) = \tau
\big(
m_e(\mathcal{R}(\tilde{S}^{i-1}), z, a_{i}^{2:3})
\big)\\
&a_{i}^4 \sim P(a_{i}^4 | z, S^{i-1}, a_{i}^{1:3})\\
\end{aligned}
\end{equation}
where $m_e(\cdot)$ is a feedforward networks. The knowledge of reaction classes can be incorporated in the same way as in Section~\ref{subsec:reaction_center}.


Given these distributions,
the distribution $p(t | z, S)$ can be parameterized according to~\cref{eq:trace_fact,eq:term_pred,eq:node_select,eq:edge_label}. 
Finally, the probability of translating the synthon $S$ to the final reactant $G$ can be computed by enumerating all possible graph transformation sequences that translate $S$ to $G$: 
\begin{equation}\label{eq:logp_gzs}
P(G|z, S) = \sum\limits_{t \in \mathcal{T}}
P(t | z, S)
\end{equation}

\subsubsection{Learning}
\label{subsubsec:step2-learning}
To learn the parameters of our variational graph translation module, we aim to maximize the log likelihood of the observed translation pair, \textit{i.e.}, $logP(G|S)$. Directly optimizing the objective involves marginalizing the latent variable $z$, which is computationally intractable. To this end, we turn to the standard amortized variational inference~\citep{kingma2014vae} by introducing an approximate posterior $q(z|G, S)$, which is modeled as a Gaussian distribution to allow effectively sampling $z$ via the reparameterization trick.
In specific, the mean and the log variance of $q(z|G, S)$ are parameterized as follows:
\begin{equation}
\begin{aligned}
& \mu = m_{\mu}(h_G \Vert h_S)\\
& \text{log}\sigma^2 = m_{\sigma}(h_G \Vert h_S)\\   
& q(z|G, S) = \mathcal{N}(z \vert \mu, \text{diag}(\sigma^2)) \\
\end{aligned}
\end{equation}
where the $h_G$ and $h_S$ are graph embeddings of $G$ and $S$ respectively, computed by the same R-GCN. $m_{\mu}(\cdot)$ and  $m_{\sigma}(\cdot)$ are feedforward networks. The evidence lower bound (ELBO) is then defined as:
\begin{equation}
\mathcal{L}_{\text{ELBO}} = \mathbb{E}_{z \sim q}[\text{log}P(G|z,S)] - \operatorname{KL}[q(z|G, S) \Vert p(z|S)]
\end{equation}
where $\operatorname{KL}[q(\cdot) \Vert p(\cdot)]$ is the Kullback-Leibler divergence between $q(\cdot)$ and $p(\cdot)$. We here take the prior $p(z|S)$ as a standard Gaussian $\mathcal{N}(z | 0, I)$. 

\textbf{Efficient Training }
The computation of $\text{log}P(G|z,S)$ (eq.~\ref{eq:logp_gzs}) is expensive as it
requires the summation over all possible graph transformation sequences that translate $S$ to $G$. 
Here we introduce two strategies that perform well empirically. 
We first show that $\text{log}P(G|z,S)$ is lower bounded by the expected log likelihood of all the trajectories $t=(a_1,\cdots,a_T)$ that translate $S$ to $G$ using Jensen{'}s inequality:
\begin{equation}
\begin{aligned}
\text{log}P(G|z, S) &= \text{log}\sum\limits_{t \in \mathcal{T}}
P(t | z, S)\\
&\geq \text{log}|t| + \frac{1}{|t|}\sum\limits_{t \in \mathcal{T}}\text{log}P(t | z, S)
\end{aligned}
\end{equation}
where $|t|$ is the number of different action traces that translate $S$ to $G$. In practice, we can throw the constant and evaluate the expectation using Monte Carlo estimation. We further adopt the breadth-first-search (BFS) node-ordering, a widely-used technique in sequential graph generation~\cite{you2018graphrnn, popova2019molecularrnn,shi*2020graphaf}, to reduce the number of valid transformation traces during sampling.

\subsubsection{Generation}\label{subsubsec:generation}

To generate a reactant graph, a natural way is to first sample a latent vector $z$ from the prior $p(z|S)$, then sample a trace of graph transformations from $p(t | z, S)$ (Section~\ref{subsubsec:generative_model}), and finally apply these transformations to the synthon $S$. However, in our proposed generative model, the probability of invalid actions will be non-zero even if the model is well-trained. As a result, any reactant molecules including invalid ones can be generated if sampling is arbitrarily long. Besides, this process also suffers from the non-trivial exposure bias problem~\cite{bengio2015scheduled}. To overcome the above obstacles 
during sampling, we design a beam search sampling process as follows.

Consider a beam search with size of $k$. 
For the graph generation in the $i^{th}$ step, we maintain a candidate set $\mathcal{S} = \{S^{i,j}\}_{j=1}^k$ with size $k$.
At the $i^{th}$ transformation step, we first calculate the probabilities of all possible actions and sort them, and then select the top $k$ ranked valid actions for each candidate graph $S^{i-1, j}$ in $\mathcal{S}$. Once this is done for $k$ graphs in $\mathcal{S}$, the top $k$ graphs among all the generated $k^2$ graphs are then selected as the candidates for the next $(i+1)^{th}$ transformation step. During this beam search, a translation branch will stop if $i$ reaches the predefined maximum transformation step or $a_i^1$ indicates a termination. In this scenario, the current graph will be added into a set $\mathcal{G}$, and the whole beam search terminates once all translation branches stop.
When the beam search finishes, the top $k$ graphs in $\mathcal{G}$,  ranked by their likelihoods, will be collected as the final predicted graphs.

\subsection{Scalability Analysis}
Both the Reaction Center Identification and the Variational Graph Translation modules in our \modelname framework bypass the deployment of reaction templates  and take advantage of the representation power of the molecular graph embeddings. As a result, the model size of \modelname scales linearly  \textit{w.r.t} the maximum number of atoms in the molecules and is invariant to the quantity of rules and reactions in the given data sets, making it highly scalable to larger data sets.

\section{Empirical Studies}

\subsection{Experiment Setup}
\textbf{Data }
We evaluate our approach on the widely used benchmark data set USPTO-50k, which contains 50k atom-mapped reactions with  10 reaction types. Following~\citep{liu2017retro_seq2seq}, we randomly select 80\% of the reactions as training set and divide the rest into validation and test sets with equal size. 

\textbf{Baselines }
We evaluate our strategy using five comparison baselines,  including  two template-free and three template-based ones. In specific, \textbf{Seq2seq}~\citep{liu2017retro_seq2seq} is a template-free approach that learns a LSTM~\citep{lstm1997} to translate the SMILES strings of target molecules to reactants. \textbf{Transformer}~\citep{pavel2019retro_transformer} is also a neural sequence-to-sequence model, but it  leverages the learning power of the Transformer~\citep{vaswani2017transformer} for better sequential modeling. \textbf{Retrosim}~\citep{Connor2017retrosim} is a data-driven method that selects template for target molecules based on similar reactions in the data set. \textbf{Neuralsym}~\citep{Marwin2017neuralsym}  employs a neural model to rank templates for target molecules.  \textbf{GLN}~\citep{Dai2019GLN} is the state-of-the-art method, which samples templates and reactants jointly from the distribution learned by a conditional graphical model.

\textbf{Evaluation Metrics }
Following the existing works~\cite{liu2017retro_seq2seq, Dai2019GLN}, we use the top-$k$ exact match accuracy as our evaluation metrics. For comparison purpose, we used $k$ with 1, 3, 5, and 10 in our experiments.  Also, the accuracy was computed by matching the canonical SMILES strings of the predicted molecules with the ground truth. 

\textbf{Implementation Details }
\modelname is implemented in Pytorch~\cite{pytorch2017automatic}. We use the open-source chemical software RDkit~\citep{RDKit} to preprocess molecules for the training and generate canonical SMILES strings for the evaluation. The R-GCN in \modelname is implemented with 4 layers and the embedding size is set as 512 for both modules.  We use latent codes of dimension $|z| = 10$.
We train our \modelname for 100 epochs with a batch size of 128 and a learning rate of 0.0001  with Adam~\citep{kingma2014adam} optimizer on a single GTX 1080Ti GPU card. The $\lambda$ is set as 20 for reaction center identification module, and the beam size is 10 during inference. 
The maximal number of transformation steps is set as 20.
We heuristically selected these values  based on the validation data set.

\begin{table}[htbp]

\centering
 \vspace{-7pt}

 \caption{Top-$k$ exact match accuracy when reaction class is given. Results of all baselines are directly taken from~\citep{Dai2019GLN}.}  
 \vspace{1mm}
 \setlength{\tabcolsep}{4mm}{
    \begin{tabular}{lcccc}
        \toprule
         \multirow{2}{*}{Methods}& \multicolumn{4}{c}{Top-$k$ accuracy \%} \\
        \cmidrule{2-5}
        & 1 & 3 & 5 & 10 \\
        \midrule
        \multicolumn{5}{c}{Template-free} \\
        \midrule
         Seq2seq & 37.4 & 52.4 & 57.0 & 61.7  \\
		 \modelname & {\bf 61.0} & {\bf 81.3} & {\bf 86.0} & {\bf 88.7}  \\
		 \midrule
		\multicolumn{5}{c}{Template-based} \\
		 \midrule
		 Retrosim & 52.9 & 73.8 & 81.2 & 88.1 \\
		 Neuralsym & 55.3 & 76.0  & 81.4 & 85.1  \\
		 GLN & {\bf 64.2} & {\bf 79.1} & {\bf 85.2} & {\bf 90.0} \\
		 \bottomrule
    \end{tabular}
    }
    \vspace{-4pt}
    \label{tab:res_class}

\end{table}

\begin{table}[!h]
\centering
 \vspace{-7pt}
 \caption{Top-$k$ exact match accuracy when reaction class is unknown. Results of all baselines are taken from~\citep{Dai2019GLN}.}  
 \vspace{1mm}
 \setlength{\tabcolsep}{4mm}{
    \begin{tabular}{lcccc}
        \toprule
         \multirow{2}{*}{Methods}& \multicolumn{4}{c}{Top-$k$ accuracy \%} \\
        \cmidrule{2-5}
         & 1 & 3 & 5 & 10 \\
        \midrule
        \multicolumn{5}{c}{Template-free} \\
        \midrule
         Transformer & 37.9 & 57.3 & 62.7 & /  \\
		 \modelname &  {\bf 48.9}&  {\bf 67.6} &  {\bf 72.5} & {\bf 75.5}  \\
		 \midrule
		\multicolumn{5}{c}{Template-based} \\
		 \midrule
		 Retrosim & 37.3 & 54.7 & 63.3 & 74.1 \\
		 Neuralsym & 44.4 & 65.3  & 72.4 & 78.9  \\
		 GLN & {\bf 52.5} & {\bf 69.0} & {\bf 75.6} & {\bf 83.7} \\
		 \bottomrule
    \end{tabular}
    }
	\vspace{-4pt}
	\label{tab:res_noclass}

\end{table}

\subsection{Predictive Performance}\label{subsec:main_res}
We evaluate the top-$k$ exact match accuracy of the proposed approach in both reaction class known and reaction class unknown settings, with results  presented in  Tables~\ref{tab:res_class} and~\ref{tab:res_noclass}, respectively. 
The sets of reactant molecules are generated via beam search (Section~\ref{subsubsec:generation}) and ranked by their log likelihoods.  

\begin{table}[htbp]
\centering
 \vspace{-5pt}
 \caption{Accuracy of the reaction center prediction module.}  
 \vspace{1mm}
    \begin{tabular}{lcccc}
        \toprule
         \multirow{2}{*}{Setting}& \multicolumn{4}{c}{Top-$k$ accuracy \%} \\
        \cmidrule{2-5}
        & 1 & 2 & 3 & 5 \\
        \midrule
		 reaction class known &  90.2 &  94.5 &  94.9 &  95.0  \\
	     reaction class unknown &  75.8 &  83.9 &  85.3 &  85.6 \\

		 \bottomrule
    \end{tabular}
	\vspace{-4pt}
\label{tab:res_reaction_center}  

\end{table}
\begin{table}[htbp]
\centering
 \vspace{-7pt}
 \caption{Accuracy of the variational graph translation module.}  
 \vspace{1mm}
    \begin{tabular}{lcccc}
        \toprule
         \multirow{2}{*}{Setting}& \multicolumn{4}{c}{Top-$k$ accuracy \%} \\
        \cmidrule{2-5}
        & 1 & 3 & 5 & 10 \\
        \midrule
		 reaction class known &  66.8 & 87.2 & 91.5 & 93.9  \\
	     reaction class unknown & 61.1  & 81.5  & 86.7  &  90.0  \\

		 \bottomrule
    \end{tabular}
	\vspace{-5pt}
\label{tab:res_vg2g}  

\end{table}

When compared with template-free approaches, results shown in Tables~\ref{tab:res_class} and~\ref{tab:res_noclass} demonstrate that the \modelname achieves competitive results on all the cases with different $k$. 
In particular, our \modelname always outperforms the template-free baselines by a large margin, with up to 63\% relative improvement in terms of the top-1 exact match accuracy when the reaction class is known (the second column in Table~\ref{tab:res_class}), and up to 29\% relative improvement when the reaction class is unknown (the second column in Table~\ref{tab:res_noclass}). 

When consider the comparison with template-based baselines, the results in Tables~\ref{tab:res_class} and~\ref{tab:res_noclass} indicate that our \modelname approaches or outperforms the state-of-the-art method GLN~\citep{Dai2019GLN}, especially when the $k$ is small. 
For example, when reaction class is given as a prior, our \modelname outperforms the GLN in terms of the top-3 and top-5 exact match accuracy.

\begin{figure*}[hbtp]
	\centering
    \includegraphics[width=.89\linewidth]{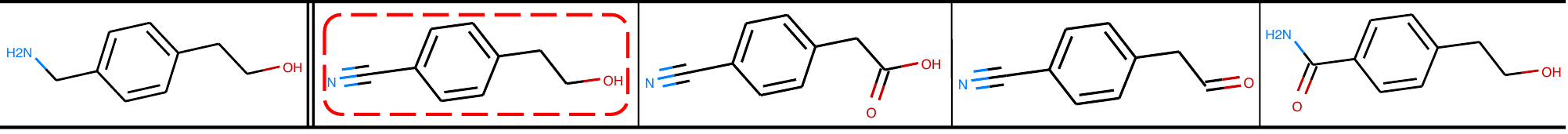}
     \vspace{-3pt}
    \caption{Visualization of the top-1 translation results (right four molecules) of a given product molecule (leftmost), which are conditioned on different latent vectors sampled from prior $\mathcal{N}(z | 0, I)$. The correct prediction provided is highlighted in red dashed box.}
    \label{fig:latent}
\end{figure*}

\subsection{Ablation Study}
To  gain  insights into the working behaviours of  G2Gs, we conduct ablation studies in this section. 

\begin{figure*}[ht]
\centering

\noindent\begin{minipage}[t]{0.48\linewidth}
\centering
\includegraphics[width=0.9\linewidth]{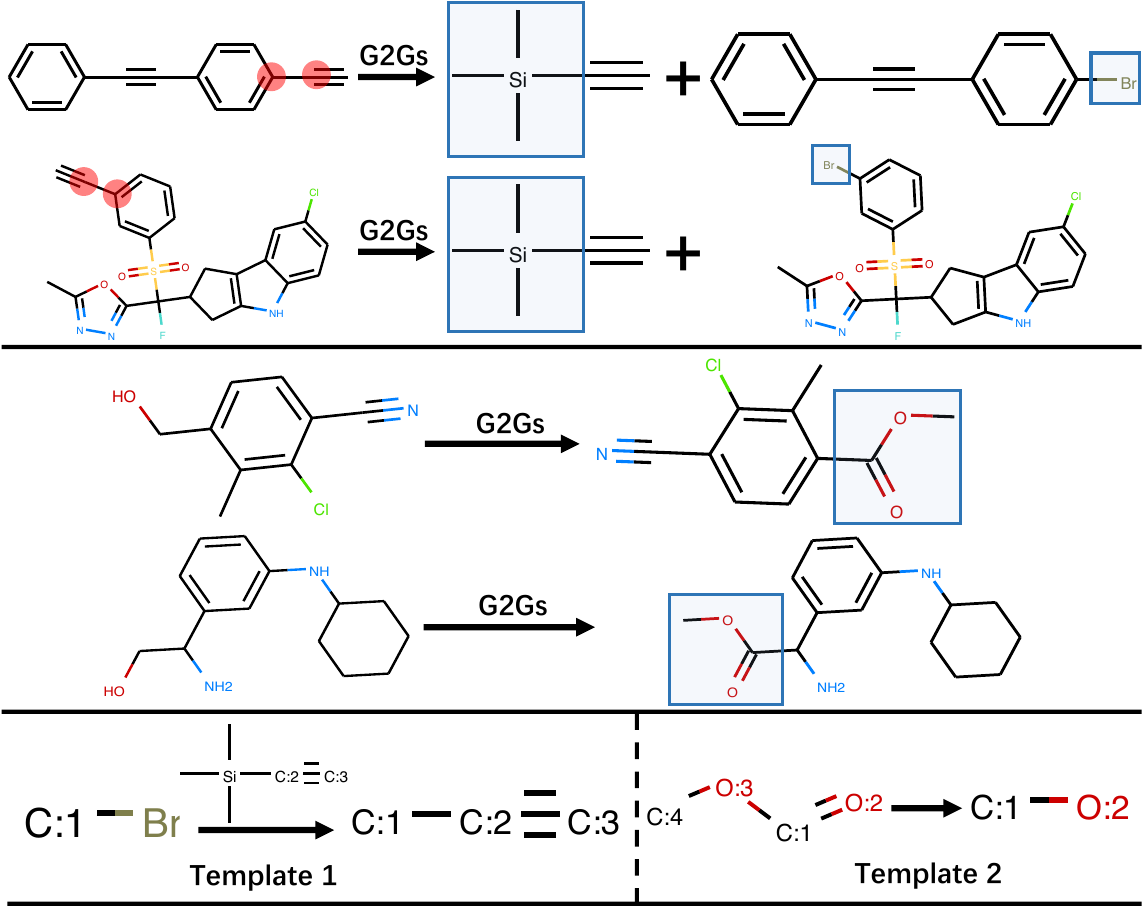}
\vspace{-2pt}
\caption{Visualization of several successful cases. Templates for these reactions are summarized at the bottom of the figure.}
\label{fig:right_case_template}
\vspace{-5pt}
\end{minipage}%
\hfill
\noindent\begin{minipage}[t]{0.48\linewidth}
\centering
\includegraphics[width=0.9\linewidth]{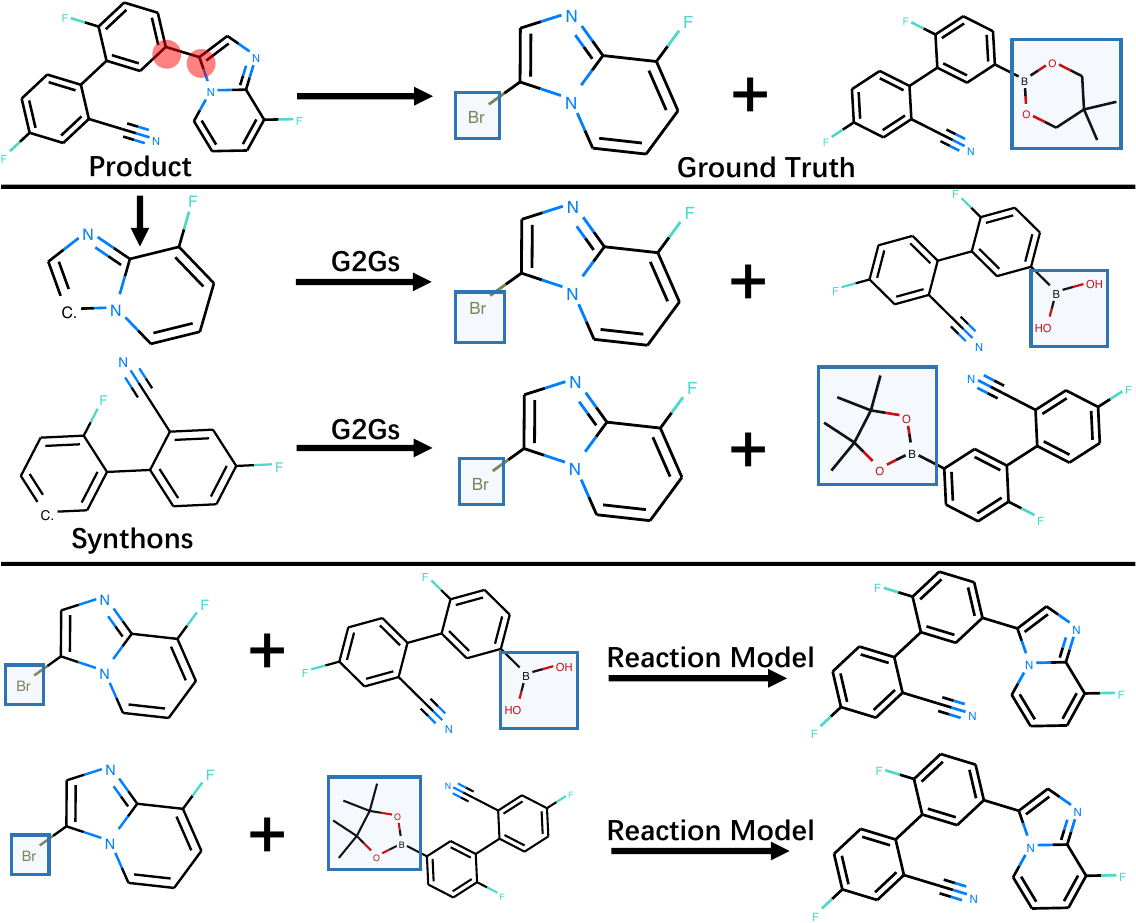}
\vspace{-2pt}
\caption{
Visualization of a mismatched case. Outcomes  of the reactions below the middle line are predicted by the forward reaction prediction model from~\citep{jin2017weisfeiler}.
}
\label{fig:wrong_case}
\vspace{-5pt}

\end{minipage}%

\centering
\end{figure*}

\textbf{Effectiveness of the Reaction Center Identification Module }
We evaluate the top-$k$ accuracy of the reaction center identification module by comparing the true reaction center against the top-$k$ atom pairs with the highest reactivity scores above a threshold. 
The results in Table~\ref{tab:res_reaction_center} indicate that when the reaction class is given as a prior, \modelname can  precisely pinpoint the reaction center in most cases even for $k=1$. When the reaction class is unknown, the performance is slightly worse than the previous case, as a target molecule can usually be synthesized in different routes depending on what reaction type a chemist chooses, and thus the model tends to make predictions with low certainty. 


\textbf{Impact of the Variational Graph Translation Module } 
To examine the  graph translation module, we first split synthons from products based on the true reaction centers (\textit{i.e.}, label matrix $Y$ in Section~\ref{subsec:reaction_center}), and then use the same strategy as in Section~\ref{subsec:main_res} to compute the top-$k$ exact match accuracies. As shown in Table~\ref{tab:res_vg2g}, the graph translation module achieves high accuracy on translating the given synthon graphs to reactant graphs, which is an important attribute to  the  superior performance of the proposed G2Gs framework. 

\textbf{Diverse Reactant Graph Generation} 
To observe the benefits brought by the latent vector used in  \modelname, Figure~\ref{fig:latent}  visualizes the top-1 translation results for a given  product molecule, which is randomly selected from the test set. In the figure, the right four molecules are generated, based on the same synthon (leftmost molecule in the figure), through randomly sampling from the prior distribution of the formed latent vector. 
Results in this figure suggest that, through sampling the latent vector formed, our method can generate diverse molecule structures. That is, sampling the learned latent vector,  an intrinsic process in  \modelname, powers our model to create diverse and valid reactant graphs, which represents an  appealing feature  for retrosynthesis prediction.

\subsection{Case Study via Visualization}

In Figure~\ref{fig:right_case_template}, we show  cases where \modelname successfully identifies the reaction centers and translates the product graph to a set of reactant graphs that match the ground truth. 
The synthesis routes shown in Figure~\ref{fig:right_case_template} can be divided into two groups, each of which corresponds to a reaction template presented at the bottom of the figure. These figures suggest that our model does learn domain knowledge from the data set. Such property of our method makes it 
an appealing solution to practical problems with limited  template knowledge.

In Figure~\ref{fig:wrong_case}, we also present a case where none of the predictions matches the ground truth.
However, we note that this does not necessarily mean that our model fails to predict a synthesis route for the target molecule. This is because a molecule can be synthesized in multiple ways and the ground truth in the data set is not the only answer. To verify  this hypothesis, we employ a forward reaction prediction model~\citep{jin2017weisfeiler} to predict the product molecules based on the reactants generated by our model. As shown at the bottom of Figure~\ref{fig:wrong_case}, the predicted product exactly matches the target molecule of the retrosynthesis problem. This confirms that  the predictions made by \modelname are indeed potentially valid.

\section{Conclusion and Outlook}

We novelly formulated retrosynthesis prediction as a graph-to-graphs translation task and proposed a template-free approach to attain the  prediction goal. In addition, 
 we devised a variational graph translation module to capture the uncertainty and to encourage diversity in the graph translation  process.  
 We also empirically verified the superior performance of our proposed method; our strategy outperformed the state-of-the-art  template-free counterpart by up to 63\% and  approached  the  performance obtained by the state-of-the-art template-based strategy, in terms of top-1 accuracy.
 Our method  excludes  domain knowledge and scales well to large datasets, making it  particularly attractive in practice.

 Our future work will include  extending our \modelname approach to embrace an end-to-end training paradigm and leveraging it to cope with  multi-step retrosynthesis tasks~\citep{corey1991logic}. 

\section*{Acknowledgements}

We thank the anonymous reviewers for their instructive feedback. This project is supported by the Natural Sciences and Engineering Research Council of Canada,
the Canada CIFAR AI Chair Program, collaboration grants between Microsoft Research and Mila, Amazon Faculty Research Award, Tencent AI Lab Rhino-Bird Gift Fund and a NRC Collaborative R\&D Project (AI4D-CORE-06).
\bibliography{reference}
\bibliographystyle{icml2020}


\end{document}